\documentclass{article}
\usepackage{spconf,amsmath,graphicx,multirow,adjustbox}
\usepackage[rgb]{xcolor}
\usepackage{flushend}
\usepackage[font=small,skip=3pt]{caption}
\def\etal{\emph{et al.}}
\usepackage{mathrsfs} 
\usepackage{calrsfs} 
\usepackage{amssymb}
\usepackage{cite}
\usepackage{enumitem}
\usepackage{afterpage}
\usepackage[subtle]{savetrees}

\title{Atmospheric Turbulence Removal with Video Sequence Deep Visual Priors}
\name{P. Hill, N. Anantrasirichai, A. Achim, and D.R. Bull }
\address{Visual Information Laboratory, University of Bristol, Bristol, UK}

\begin{document}
\ninept
\maketitle
\begin{abstract}

\noindent Atmospheric turbulence poses a challenge for the interpretation and visual perception of visual imagery due to its distortion effects. Model-based approaches have been used to address this, but such methods often suffer from artefacts associated with moving content. Conversely, deep learning based methods are dependent on large and diverse datasets that may not effectively represent any specific content. In this paper, we address these problems with a self-supervised learning method that does not require ground truth. The proposed method is not dependent on any dataset outside of the single data sequence being processed but is also able to improve the quality of any input raw sequences or pre-processed sequences.  Specifically, our method is based on an accelerated Deep Image Prior (DIP), but integrates temporal information using pixel shuffling and a temporal sliding window. This efficiently learns spatio-temporal priors leading to a system that effectively mitigates atmospheric turbulence distortions. The experiments show that our method improves visual quality results qualitatively and quantitatively. 
\end{abstract}
\section{Introduction}
\label{sec:intro}

Image and video acquisition can suffer from visual degradation caused by visual distortions associated with atmospheric conditions.
Visual distortions often arise when there is a significant atmospheric temperature gradient between the ground and the surrounding air. As a result, the layers of air ascend rapidly, leading to spatially diverse alterations in the index of refraction along the optical path. This phenomenon is commonly observed as a combination of blurriness, ripples, and fluctuations in the intensity of the scene. Instances of this effect can be found in areas associated with hot surfaces (such as roads and runways) and deserts, as well as in close proximity to heated human-made objects like aircraft jet exhausts. Particularly in hot environments near the ground, this poses a significant challenge and can compound with other detrimental factors in long-range surveillance applications, such as fog or haze, which also diminish contrast and video quality. Atmospheric distortions primarily stem from elevated temperatures and long capture distances, with additional influential factors including altitude, wind speed, humidity, and pollution. 

The modelling of atmospheric turbulence within recorded videos presents a complex and time-consuming task due to the ill-posed nature of the problem, characterised by spatio-temporally variant distortions.
Moreover, this phenomenon adversely affects contrast, sharpness, and the ability to discern distant objects, especially in surveillance applications. It also impacts the effectiveness of visual analytics operations such as detection, classification and tracking.  
Alongside generic inverse problems, atmospheric turbulence can be represented through the following relationship:
\[
\boldsymbol{y} = D\boldsymbol{x} + \boldsymbol{n},
\]
where $\boldsymbol{x}$ and $\boldsymbol{y}$ are the undistorted and observed images or videos respectively. The transform $D$ represents an unknown spatio-temporal distortion caused by the system, while $\boldsymbol{n}$ represents the noise present in the observation. Despite this representation being simple it is inherently irreversible, leading to imperfect solutions in practical applications.

Conventional approaches attempt to invert this system through modelling $D$ as a point spread function (PSF) and employing blind deconvolution techniques with an iterative process to estimate the ideal image $\boldsymbol{x}$ \cite{harmeling2009online}. Conversely, image fusion techniques offer an alternative solution by selectively combining frames to reconstruct a clearer image, utilising only the high-quality information available~\cite{anantrasirichai2012mitigating}.  Several authors (e.g.~\cite{mao2022single,chimitt2022real,Mao:accelaring:2021}) have chosen to break down the transform $D$ into separate geometric distortion and blur operations. However, despite the conceptual value of this approach, it is impossible to accurately separate these individual transformations, and their closed-form separation overlooks additional multiplicative distortion effects resulting from phenomena like pollution, fog, and haze.

Model-based methods encounter two significant challenges. Firstly, they exhibit high computational complexity, making real-time implementation nearly impossible. Secondly, the combination of multiple images can introduce artefacts from moving objects due to imperfect alignment~\cite{anantrasirichai2018atmospheric}.  However, model based methods such as CLEAR~\cite{anantrasirichai2018atmospheric} still provide competitive and state of the art results even when compared with more recently developed deep learning / trained methods.

The application of deep learning has proven to be highly effective in the identification of patterns, data analysis, and prediction of future events. These capabilities have resulted in the widespread adoption of deep learning-based techniques in the field of image and video processing. However, in the domain of atmospheric turbulence removal, deep learning is still in its early stages, with the majority of proposed methods relying on Convolutional Neural Networks (CNNs)~\cite{Rai:learning:2020, Gao:Atmospheric:2019,anantrasirichai2023atmospheric, Vint:analysis:2020}, trained with either synthetic distortions, which are generally too simplified, or pseudo ground truth, which is obviously not perfectly clean.

An initial deep learning-based method was introduced by Gao \etal~\cite{Gao:Atmospheric:2019}.  This work assumes that the spatial displacement between frames caused by atmospheric turbulence follows a Gaussian distribution. The method utilised the CNN based Gaussian denoiser, DnCNN~\cite{Zhang:dncnn:2017}, architecture. Mao \etal \cite{Mao:accelaring:2021} have subsequently, proposed a similar method but instead employed the UNet architecture (originally designed for medical image segmentation \cite{Ronneberger:Unet:2015}). 
Vint~\etal \cite{Vint:analysis:2020} conducted a study on the performance of various state-of-the-art architectures, originally developed for denoising, deblurring, and super-resolution, in mitigating the effects of atmospheric turbulence. Their findings, as reported, show great promise. However, it is important to note that their investigation was limited to synthetic static scenes.

In this paper, we address the problems stated above with the following contributions:

\begin{itemize}
    \item The definition of an effective architecture for processing image sequences motivated by the Deep Image Prior method (DIP~\cite{ulyanov2018deep})
    \item Optimisation of a Deep Image Prior based method with Deep Random Projection (DRP~\cite{li2023deep}) and Early Stopping (ES~\cite{wang2021early}) together with latent variable prediction for further acceleration
    \item The evaluation of DIP methods using existing No Reference (NR) and novel spatio-temporal metrics 
    \item Generating a turbulence mitigation system that is able to reduce spatial and temporal artefacts together with improving image and sequence quality quantitatively and qualitatively.
\end{itemize}

\section{Proposed methodology}
\subsection{Deep Image Prior (DIP)}
DIP is based on the concept of representing an observed signal (in most cases a visual object) $\boldsymbol{x}$, using a structured Deep Neural Network (DNN)~\cite{2017Deep}. This is achieved by obtaining $\boldsymbol{x}$ through the function $\boldsymbol{x}=G_\theta(\boldsymbol{z})$, where $G_\theta$ represents the DNN and $\boldsymbol{z}$ is a latent input (a randomly chosen seed).

An inverse problem of form $\boldsymbol{y}\approx f(\boldsymbol{x})$ can be characterized by $f$: the observation process, together with a minimisation process (inspired by a Maximum-a Posteriori (MaP) estimation process) with a regularisation term $R(\boldsymbol{x})$ thus leading to the form:

\[
\underset{\boldsymbol{x}}{\text{min }} \underbrace{\ell(\boldsymbol{y},f(\boldsymbol{x}))}_{\text{data fitting loss}} + \underbrace{\lambda R(\boldsymbol{x})}_{\text{regularistation term}}
\]

\noindent where $\lambda$ is a scalar regularisation parameter and $\ell$ is the loss function. For any 2D image $\boldsymbol{x} \in \mathbb{R}^{h\times w}$.
 Plugging in the  representation of a signal output of a model $G_\theta$ given an input latent variable $\boldsymbol{z}$ (i.e. $\boldsymbol{x}=G_\theta(\boldsymbol{z}))$ we obtain:

\[
\underset{\boldsymbol{\theta}}{\text{min }} \ell(\boldsymbol{y},f \circ G_\theta(\boldsymbol{z})) + \lambda R \circ G_\theta(\mathbf{\boldsymbol{z}}).
\]

\subsubsection{Overfitting and Early Stopping (ES)}
Compared to the input signal, the model $G_{\theta}$ could be significantly over-parametrised.  It is therefore theoretically possible for $f \circ G_\theta(\boldsymbol{z})$ to very accurately match the input $\boldsymbol{y}$, particularly when no additional regularisation is applied. DIP depends on the ``early-learning-then-overfitting" phenomenon. During the learning process, the model primarily learns the desired visual content before fitting the noise \cite{li2021self,wang2021early}. This phenomenon is believed to be a combined implicit regularisation effect of over-parametrisation and convolutional structures in $G_\theta$.  Therefore, if the point of peak performance can be identified and the fitting process halted at that point (by employing an appropriate early stopping technique), a reliable estimate for $\boldsymbol{x}$ can be obtained.  Numerous ES methods exist~\cite{li2021self,wang2021early}.  However, we adopt the ES-WMV Windowed Moving Variance (WMV) developed by Wang \etal~\cite{wang2021early} as it was the best performing and most appropriately defined for our application found in the literature.

\subsubsection{DIP Acceleration}
To obtain the most flexible training where there is no other data required other than the input, a naive DIP method fixes the latent input ($\boldsymbol{z}$) and performs Stochastic Gradient Descent (SGD) on the network weights $\boldsymbol{\theta}$ through the minimisation of the loss between $G_{\boldsymbol{\theta}}$ and $\boldsymbol{y}$. This is computationally very expensive due not only to the over-parameterised nature of $G_{\boldsymbol{\theta}}$ but also to a lack of regularisation.  We firstly accelerate the DIP by adopting the optimisation strategies defined as Deep Random Projection (DRP) by Li \etal~\cite{li2023deep} together with:
\begin{itemize}
    \item Adopting a reduced size hour-glass network (as defined within~\cite{li2023deep}). 
    \item Freezing all weights $\theta$ within the $G_{\boldsymbol{\theta}}$ except the weights associated with Batch Normalisation ($\theta_{BN}$).
    \item Unfreezing the latent input variable $\boldsymbol{z}$ and learning updates for $\boldsymbol{z}$ in the SGD process.
    \item Adopting a weighted and explicit regularisation term based on total variation, defined as in (\ref{eq:TV}).
\end{itemize}

\begin{equation}
   TV(\boldsymbol{x}) = \sum_{i,j}|\boldsymbol{x}_{i+1,j}-\boldsymbol{x}_{i,j}|+|\boldsymbol{x}_{i,j+1}-\boldsymbol{x}_{i,j}| 
\label{eq:TV} 
\end{equation}
These factors are combined to form the following update rule:
\[
\underset{\boldsymbol{z}, \theta_{BN}}{\text{min }} \ell(\boldsymbol{y} - f \circ G_\theta(\boldsymbol{z})) + \lambda TV( G_\theta(\boldsymbol{z})).
\]

\begin{figure}
    \centering
    \includegraphics[width=0.75\linewidth]{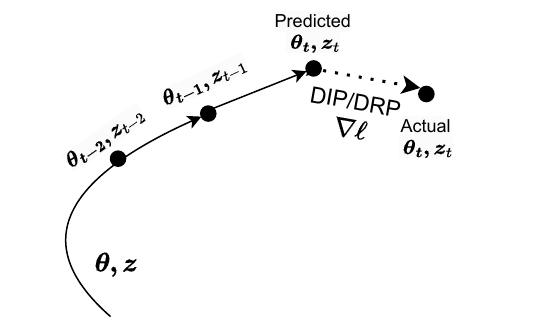}
    \caption{Latent Variable / Batch Normalisation weights prediction\, \{$\theta_{BN}, \boldsymbol{z}$\}} 
    \label{fig:z}
\end{figure}

\noindent As the majority of the weights within the model are frozen, the dependency on the output becomes more focused on the latent variable input.  The semantic relationship between close locations in latent variable space of GANs~\cite{karras2020analyzing} has been recently recognised. Initialising the learnt variables\, \{$\theta, \boldsymbol{z}$\} with random values leads to a significant amount of time being taken to converge to a good approximation of the input sequence window in the SGD process. To address this, we have chosen to re-initialise the next temporal step with the previous values of\, \{$\theta, \boldsymbol{z}$\}. However, we can improve upon this approach by using linear prediction of\, \{$\theta, \boldsymbol{z}$\} from its previous values, as illustrated in Figure \ref{fig:z}.

\begin{figure}
    \centering
    \includegraphics[trim={5cm 17cm 5cm 0cm},clip,width=1.03\linewidth]{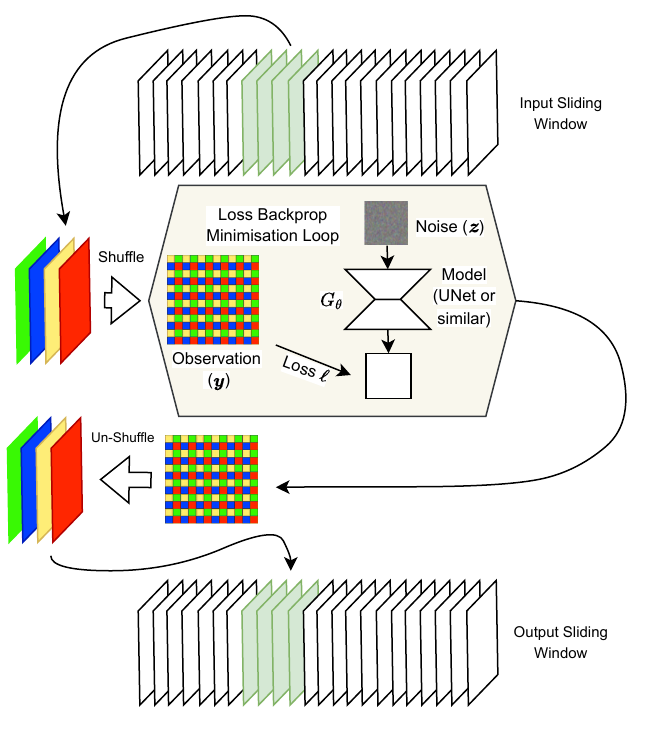}
    \caption{Architecture of our developed image sequence atmospheric mitigation method}
    \label{fig:arch}
\end{figure}

\subsection{Architecture and implementation}
Figure~\ref{fig:arch} shows the architecture of our sequence based DIP / DRP system. This architecture uses a sliding window to assemble a group of images to be processed as a group.  Instead of implementing a 3D spatio-temporal version of the DIP / DRP method, we have adopted the shuffling mechanism which interlaces the input frames into a mosaic as shown in the figure~\cite{Shi:real:2016}.  The shuffling mechanism has been found to give significant advantages over processing in higher dimensions while being able to characterise and process the dependencies between temporally related samples / pixels~\cite{Shi:real:2016}.  The DIP / DRP method requires that the input and outputs are of the same dimension.  Therefore each 3D spatio-temporal volume is processed one block of frames at a time with a range of frames being placed into the output sequence for each processing unit.  It was anticipated that this would lead to temporal artefacts.  However, this was not found to be the case.
The parameters for our implementation were as follows.
\begin{itemize}
    \item Early Stopping Parameters  (as defined by the WMV method in \cite{li2021self}). These values were significantly less than in the paper due to the acceleration given by latent variable prediction: Patience = 50, Max Epoch = 200, Patience Start = 50, Alpha = 0.1.

    \item Architecture Parameters: A small hour-glass network is employed (as defined within~\cite{li2023deep}). Each colour image is transformed into a monochrome single channel image with the colour data only being placed back on the image for visualisation after processing. The quantitative metric values were only calculated for the single channel monochrome images. Number of frames per block in sliding window is 5. Acceleration uses latent variable / batch normalisation weight prediction in image sequence based DIP. \,SGD for the first two frame blocks is directly from randomly initialised \{$\theta_{t=0},\boldsymbol{z}_{t=0}$\} and \{$\theta_{t=1},\boldsymbol{z}_{t=1}$\}. Prediction of\, \{$\theta_{t>=2},\boldsymbol{z}_{t>=2}$\} is from the two previous value of\, \{$\theta,\boldsymbol{z}$\}.
    \item Regularisation: $\lambda$ = 0.1

\end{itemize}

\begin{figure*}
\centering
\caption*{\textbf{Table 1}. Results: BIQI and Variance results for input together with results for CLEAR, our method and combined our and CLEAR methods (first 5 frames of each sequence).  BIQI results show the mean and standard deviations for each set of frames.  The higher the BIQI value the lower the quality.}
\label{tab:results1}
\begin{tabular}{llll}
\hline
\hline
Sequence   & Pre-Processing & Background-Var $\downarrow$ & BIQI~\cite{Moorthy:twostep:2010} $\downarrow$  \\
\hline
\hline
\multirow{4}{*}{Van~\cite{anantrasirichai2018atmospheric}}       &  Input         &   19.622 &   36.302 $\pm 2.438\sigma$  \\
&  Input + CLEAR           &  1.855         &   34.240 $\pm 0.644\sigma$   \\
&  Input + Ours            &  15.803        &   34.371 $\pm 3.729\sigma$   \\
&  Input + CLEAR + Ours    &  \textbf{1.171}         &   \textbf{29.911 $\pm 0.543\sigma$}   \\
\hline
\multirow{4}{*}{Moving Car~\cite{anantrasirichai2018atmospheric}}&  Input         &   18.735  &  43.124 $\pm 0.631\sigma$ \\
&  Input + CLEAR           &  8.648         &   43.000 $\pm 2.302\sigma$   \\
&  Input + Ours            &  13.783        &   38.850 $\pm 0.570 \sigma$  \\
&  Input + CLEAR + Ours    &  \textbf{6.797}         &   \textbf{38.000 $\pm 2.232\sigma$}   \\
\hline
\multirow{4}{*}{VayTek~\cite{anantrasirichai2018atmospheric}}&  Input         &   265.385  &  38.361 $\pm 1.122\sigma$ \\
&  Input + CLEAR           &  1.970         &   36.464 $\pm 0.372\sigma$   \\
&  Input + Ours            &  215.852       &   36.054$\pm 0.615\sigma$  \\
&  Input + CLEAR + Ours    &  \textbf{1.262}         &   \textbf{33.014 $\pm 0.199\sigma$}   \\
\hline
\multirow{4}{*}{Train~\cite{Gao:Atmospheric:2019}}&  Input         &   0.743  &  40.060 $\pm 2.452\sigma$ \\
&  Input + CLEAR           &  0.863         &   24.125 $\pm 2.248\sigma$   \\
&  Input + Ours            &  0.353       &   39.933$\pm  2.444\sigma$  \\
&  Input + CLEAR + Ours    &  \textbf{0.341}         &   \textbf{23.229 $\pm 2.284\sigma$}   \\
\hline
\multirow{4}{*}{Shore~\cite{Gao:Atmospheric:2019}}&  Input         &   5.195  &  38.516 $\pm 6.258\sigma$ \\
&  Input + CLEAR           &  1.426         &   34.811 $\pm 3.321\sigma$   \\
&  Input + Ours            &  3.928       &   39.933$\pm  2.444\sigma$  \\
&  Input + CLEAR + Ours    &  \textbf{1.189}         &   \textbf{21.544 $\pm 0.392\sigma$}   \\
\hline
\hline
\vspace{0.5cm}
\end{tabular}
    \centering
    
    \hspace{-1.1cm}
    \includegraphics[width=0.7\linewidth]{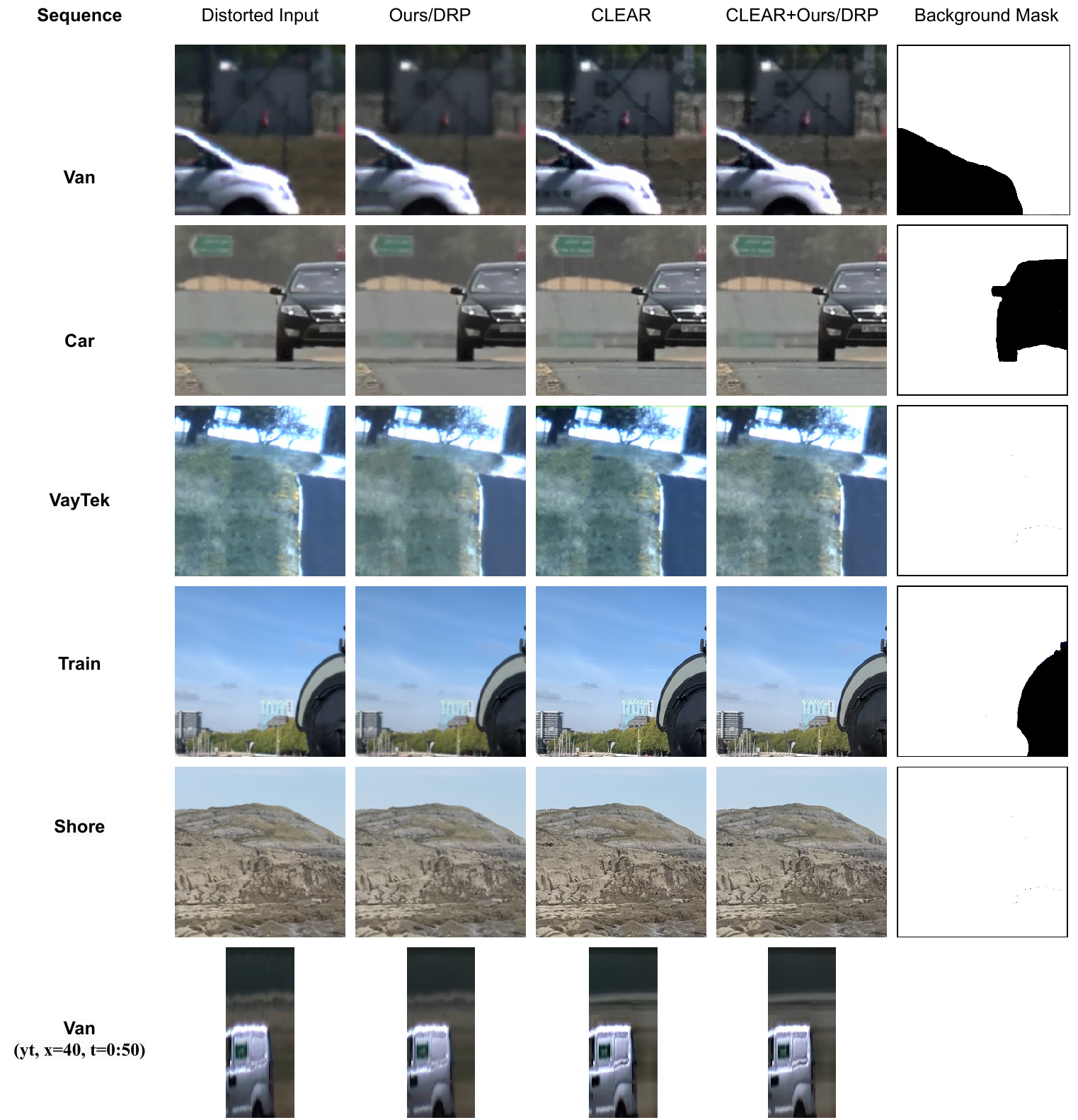}
    \caption{Frame 0 from five sequences \{van~\cite{anantrasirichai2018atmospheric}, car~\cite{anantrasirichai2018atmospheric}, Vaytek~\cite{anantrasirichai2018atmospheric}, train~\cite{Gao:Atmospheric:2019}, shore~\cite{Gao:Atmospheric:2019}\}. The figure shows the distorted input (leftmost image) together with results for CLEAR~\cite{anantrasirichai2012mitigating}, our method and combined our and CLEAR methods.  The bottom figure shows the $yt$ plane of the van sequence showing the temporal variations with a fixed $x$-axis.}
    \label{fig:results_ims}
\end{figure*}

\section{Experiments and discussion}
\subsection{Metrics}
\label{sec:metric}
Since there is no ground truth associated with any of the dataset sequences, it is not possible to use reference-based evaluation methods. Numerous Non-Reference (NR) image-based quality metrics are available~\cite{sheikh2005no, Moorthy:twostep:2010, Gabarda:blind:2007}. We have opted for BIQI (Blind Image Quality Indices)~\cite{Moorthy:twostep:2010} due to its compatibility with the turbulence mitigation problem \cite{6471221} together with its availability.  It should be noted the BIQI values vary from in the range 0-100 where higher values indicate less quality.

There have not been any effective metrics defined to evaluate the quality of turbulence mitigation methods for sequences (rather than single images).  In order to  evaluate our method for non-static scenes, we have evaluated processing outputs by measuring the variance of the background pixels.  The background pixels are masked using a hand drawn mask for each image within each sequence (see figure~\ref{fig:results_ims}).  The variance is measured along the temporal axis (where background pixels exist) and averaged across the background for the entire considered sequence.  This measure is labelled as \textbf{Background-Var} in results Table 1.  The justification for using this metric is that background areas for static camera based sequences should not move and therefore have low variance (all the considered sequences have static cameras and contain moving objects).  It is also justified by the fact that the high performing model based system CLEAR~\cite{6471221} often generates artefacts associated with the boundary of moving objects derived from the large support of the complex wavelets used within the CLEAR system.  The variations of these edge artefacts will be picked up by background variance.

\subsection{Results and discussion}
Table 1 shows results for five sequences\, \{van~\cite{anantrasirichai2018atmospheric}, car~\cite{anantrasirichai2018atmospheric}, Vaytek~\cite{anantrasirichai2018atmospheric}, train~\cite{Gao:Atmospheric:2019}, shore~\cite{Gao:Atmospheric:2019}\}.  These results are also illustrated in figure~\ref{fig:results_ims}.  The table shows that the no-reference quality of the output (measured by decreasing BIQI metric values) increases in the following order:

\begin{itemize}
    \item The input distorted sequence (lowest NR quality, highest BIQI value: highest background variance)
    \item Our method (Accelerated DIP using DRP with ES)
    \item CLEAR~\cite{anantrasirichai2012mitigating}
    \item Our method (Accelerated DIP using DRP with ES) + CLEAR~\cite{anantrasirichai2012mitigating} (highest NR quality, lowest BIQI value: lowest background variance)
\end{itemize}

\noindent The results in Table 1 indicate that the quality of all the output sequences are higher than the input (i.e.\ the input sequences have the highest BIQI values). This table also shows that the highest quality outputs are obtained through the post processing of the CLEAR output with our method.  The background-variance values shows that the background of each sequence is more static when using our method on the input and CLEAR based sequences.  Also, on visual inspection, the moving object boundary based artefacts created by the CLEAR method are significantly reduced by our method.  Although our method is not able to effectively fuse and sharpen the content of turbulent sequences to the extent of the CLEAR method, it shows significant utility in not only reducing spatial and temporal distortions, but also reducing the effect of moving object artefacts generated by state of the art model based methods such as CLEAR.  This is also achieved with self training.  This will have great facility where it is impossible to generate any ground truth data for training any effective deep learning system.  This is often the case for turbulence data where ground truth is only available for contrived and indoor scenes.  Other turbulence mitigation methods were considered for comparison, however the majority of recent methods (e.g.~\cite{Mao:accelaring:2021,Cheng:Restoration:2023,mao2022single,Vint:analysis:2020,nair2023ddpm,Jin:Neutralizing:2021}) use some trained component and are therefore difficult to fairly compare.

\section{Conclusion}
Our proposed method combines previously developed image based Deep Image Prior (DIP)~\cite{ulyanov2018deep} systems with: Deep Random Projection (DRP), Early Stopping (ES) and Pixel Shuffling to form an effective self-supervised learning turbulence mitigation system. 
We have further accelerated our system by predicting an optimal latent variable input for each new temporal window from previously learnt latent variables.  Our method not only gives an optimised framework for using a Deep Visual Prior for image sequence processing, but also is able to generalise to any sequence given that it is not dependant on any data outside of the sequence being processed (this remains true when our method is combined with CLEAR which is itself not dependant on training data).

Our method shows superior results quantitatively and qualitatively.   Quantitative results show that our method improves the image quality (using the ND image quality metric BIQI) and reduces the background variance of both the distorted input or pre-processed CLEAR output sequences.  Furthermore, qualitatively, our method reduces the artefacts associated with CLEAR while also improving the visual quality of all outputs together with reducing spatial variance.
\bibliographystyle{IEEEbib}

\bibliography{DIP_Fusion, refs}

\end{document}